\documentclass[10pt,twocolumn,letterpaper]{article}

\usepackage{wacv}
\usepackage{times}
\usepackage{epsfig}
\usepackage{graphicx}
\usepackage{amsmath}
\usepackage{amssymb}
\usepackage{booktabs}
\usepackage{hyperref}
\hypersetup{
	colorlinks = true, 
	urlcolor = magenta, 
	linkcolor = red, 
	citecolor = blue}

\usepackage{enumitem}
\usepackage{romannum}
\usepackage{booktabs}
\usepackage{amsmath}
\usepackage{cuted}
\usepackage{capt-of}

\usepackage{array}
\newcolumntype{P}[1]{>{\centering\arraybackslash}p{#1}}

\usepackage{pythonhighlight}

\def\wacvPaperID{1787} 

\wacvalgorithmstrack   

\wacvfinalcopy 

\begin{document}

\title{Treating Motion as Option to Reduce Motion Dependency\\in Unsupervised Video Object Segmentation}
\author{Suhwan Cho$^1$\quad Minhyeok Lee$^1$\quad Seunghoon Lee$^1$\quad Chaewon Park$^1$\\
Donghyeong Kim$^1$\quad Sangyoun Lee$^{1,2}$\vspace{0.5cm}\\
$^1$~~Yonsei University\\
$^2$~~Korea Institute of Science and Technology (KIST)}
\maketitle

\pagenumbering{gobble} 

\begin{strip}
    \vspace{-1.5cm}
    \centering
    \includegraphics[width=1.0\textwidth]{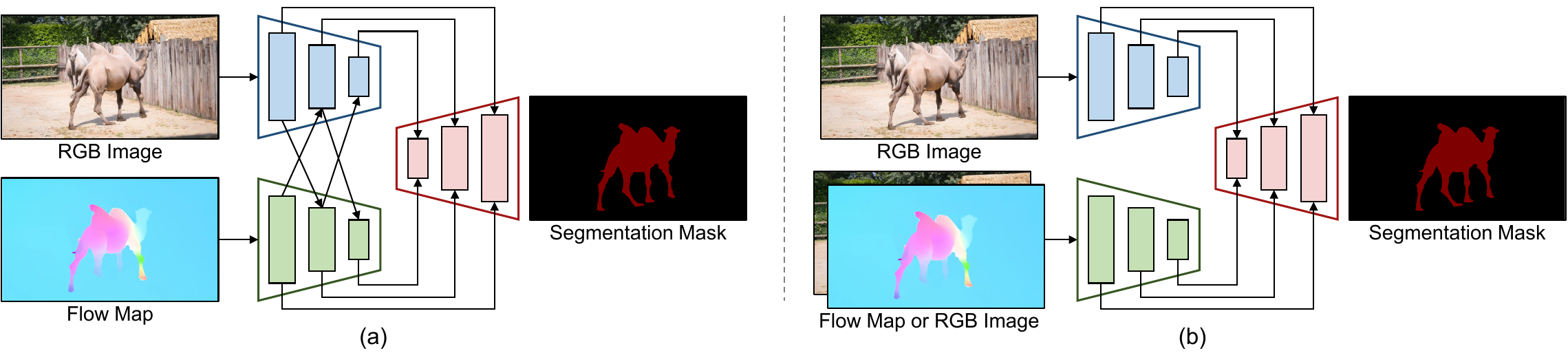}
    \captionof{figure}{Visualized comparison of (a) a conventional two-stream VOS network and (b) our proposed motion-as-option network.\label{figure1}}
\end{strip}

\begin{abstract}
Unsupervised video object segmentation (VOS) aims to detect the most salient object in a video sequence at the pixel level. In unsupervised VOS, most state-of-the-art methods leverage motion cues obtained from optical flow maps in addition to appearance cues to exploit the property that salient objects usually have distinctive movements compared to the background. However, as they are overly dependent on motion cues, which may be unreliable in some cases, they cannot achieve stable prediction. To reduce this motion dependency of existing two-stream VOS methods, we propose a novel motion-as-option network that optionally utilizes motion cues. Additionally, to fully exploit the property of the proposed network that motion is not always required, we introduce a collaborative network learning strategy. On all the public benchmark datasets, our proposed network affords state-of-the-art performance with real-time inference speed. Code and models are available at \url{https://github.com/suhwan-cho/TMO}.
\end{abstract}

\section{Introduction}
Video object segmentation (VOS) is a fundamental computer vision task that aims to detect an object or objects in a given video sequence at the pixel level. Owing to its powerful applicability in real-world applications, such as robotics, video editing, and autonomous driving, it is widely used in many vision systems. Based on the guidance on the object to segment, VOS can be divided into many subcategories, such as semi-supervised VOS (initial mask guidance), unsupervised VOS (no guidance), weakly-supervised VOS (initial box guidance), interactive VOS (human guidance), and referring VOS (language guidance). This study focuses on unsupervised VOS, which is also known as zero-shot VOS, where no manual annotation about the target object to be segmented is provided. As no clear guidance is provided about the object, the aim is to automatically define the salient object of a given video sequence and consistently segment that object for the entire frame.

Inspired by the observation that a salient object usually has distinctive movements compared to the background, recent approaches for unsupervised VOS utilize motion cues obtained from optical flow maps as well as appearance cues obtained from RGB images. Appearance and motion cues are fused during the feature embedding process to provide mutual guidance, as shown in Figure~\ref{figure1}~(a). MATNet~\cite{MATNet} comprises a deeply-interleaved encoder architecture for hierarchical interaction between the object appearance and motion. FSNet~\cite{FSNet} comprises a relational cross-attention module for mutual correction and bidirectional interaction. Additionally, AMC-Net~\cite{AMC-Net} has a multi-modality co-attention gate to integrate cross-modality features into unified feature representations. Although these methods achieve remarkable performance, they are susceptible to low-quality optical flow maps as they are strongly dependent on motion cues. 

To overcome this limitation, we design a novel network that operates regardless of motion availability, termed as a motion-as-option network, as shown in Figure~\ref{figure1}~(b). The proposed network is based on simple encoder--decoder architecture. The encoder extracts semantic features from an RGB image and an optical flow map, while the decoder generates an object segmentation mask by fusing and decoding those features. In contrast to the existing two-stream VOS methods such as MATNet, FSNet, and AMC-Net, we separately encode the respective cues and fuse them via simple summation after the feature embedding process. Furthermore, the motion stream is adaptively turned on or off, making it less dependent on motion cues. If the motion stream is turned on, motion cues are added to the appearance cues to construct the final cues, and if it is turned off, only the appearance cues are used to construct the final cues (RGB images are used as input of the motion encoder). The proposed motion-as-option network has two main advantages over the existing two-stream methods: 1) it is robust against low-quality optical flow maps as the network is learned while treating motion as option; and 2) it can run without optical flow maps during inference, which greatly increases its applicability and usability. 

To fully exploit the properties of the proposed motion-as-option network, we also propose a collaborative network learning strategy. As motion cues are optionally adopted in the proposed network, both training samples with and without optical flow maps need to be provided in the training stage to meet our network design goal. A straightforward approach to obtain these training samples is intentionally and randomly discarding optical flow maps in VOS training samples. However, to better exploit the advantage that optical flow maps are not always needed, we additionally adopt salient object detection (SOD) training samples to provide large-scale data. When the motion stream is turned on, VOS training samples are used and optical flow maps are fed into the motion encoder as input. When the motion stream is turned off, SOD training samples are used and RGB images are fed into the motion encoder as input.

We validate our proposed approach on the public benchmark datasets for unsupervised VOS, DAVIS 2016~\cite{DAVIS} validation set, FBMS~\cite{FBMS} test set, and YouTube-Objects~\cite{YTOBJ} dataset, by quantitatively and qualitatively comparing it to other state-of-the-art methods. On all the benchmark datasets, our proposed method outperforms the existing methods while maintaining an exceptionally-fast inference speed of 40+ fps on a single GeForce RTX 2080 Ti GPU. We believe that our simple, fast, and strong solution takes a meaningful step toward efficient and applicable VOS, and can serve as a solid baseline for future research.

Our main contributions can be summarized as follows:
\begin{itemize}[leftmargin=0.2in]
	\item We introduce a novel motion-as-option network that optionally leverages motion cues and a collaborative network learning strategy that maximizes the advantage of the proposed network. 
	
	\item The proposed motion-as-option network performs strongly against low-quality optical flow maps and can even run without optical flow maps. 

    \item On public benchmark datasets, the proposed network achieves state-of-the-art performance, with a $\mathcal{G}$ score of 86.1\% on the DAVIS 2016 validation set, a $\mathcal{J}$ score of 79.9\% on the FBMS test set, and a $\mathcal{J}$ score of 71.5\% on the YouTube-Objects dataset.
    
    \item In contrast to other recent methods that are designed with complex architectures, our method is designed with a simple encoder--decoder architecture, which enables a real-time inference speed of 40+ fps.
\end{itemize}

\section{Related Work}
\noindent\textbf{Temporal coherence.} A key property of videos is that different frames of the same video sequence share similar content that are highly related to each other. In unsupervised VOS, some methods have exploited this locality of a video, i.e., the observation that salient objects appear in every frame of a video. COSNet~\cite{COSNet} emphasizes the inherent correlation among video frames from a holistic viewpoint. It adopts a global co-attention mechanism to capture the frequently-reappearing salient object in a video. AGNN~\cite{AGNN} solves the problem of the reappearance of the salient object by regarding each frame as node and the relations between the frames as edges. Through iterative information fusion over the frames, a complete understanding of the video content can be obtained. DFNet~\cite{DFNet} obtains the inherent long-term correlation between different frames of a video by learning discriminative representation from a global perspective. AD-Net~\cite{AD-Net} and F2Net~\cite{F2Net} reconsider long-term temporal dependencies in a video by establishing dense correspondences between pixel embeddings of the reference and query frames. As these methods require multiple frames for calculating coherence, each separate frame in a video cannot be independently inferred.

\begin{figure*}[t]
	\centering
	\includegraphics[width=1\linewidth]{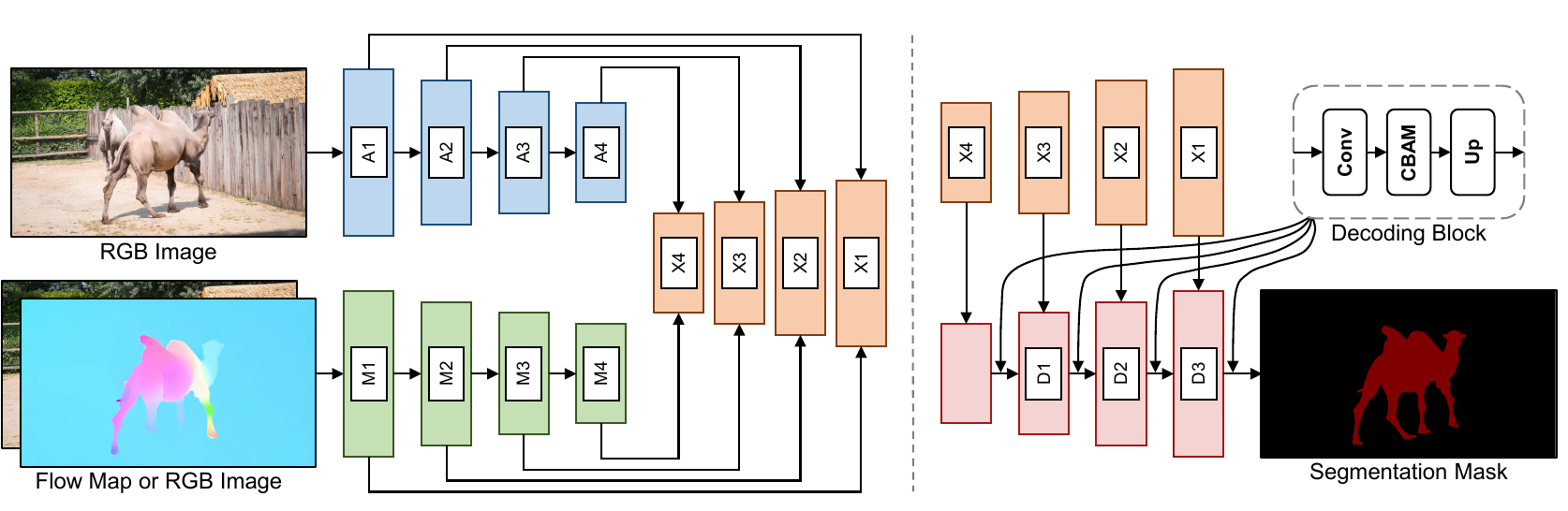}
	\caption{Architecture of our proposed network. Appearance and motion features are separately extracted from an RGB image and either an RGB image or an optical flow map, respectively. If the motion stream is turned on, both appearance and motion cues are leveraged, whereas if the motion stream is turned off, only appearance cues are leveraged. The fused features after the appearance and motion embedding are gradually decoded by the decoder to predict the final binary segmentation mask.}
	\label{figure2}
\end{figure*}

\vspace{1mm}
\noindent\textbf{Motion information.} To leverage the property that salient objects usually have distinctive motion that can be distinguished from the background, some methods have exploited the short-term motion information obtained from pre-trained optical flow estimation models. MATNet~\cite{MATNet} has a two-stream encoder that employs RGB images and optical flow maps to separately treat appearance and motion, which was performed for the first time. RTNet~\cite{RTNet} focuses on the problem that motion information is sometimes too noisy, which results in misguidance about the object. To solve this, a reciprocal transformation network is proposed to correlate the intra-frame contrast, motion cues, and temporal coherence of recurring objects. FSNet~\cite{FSNet} introduces a full-duplex strategy to better exchange useful cues between appearance and motion information. It comprises a relational cross-attention module to realize bidirectional message propagation across embedding subspaces. TransportNet~\cite{TransportNet} establishes the correspondence between appearance and motion cues while suppressing the distracting noises via optimal structural matching using Sinkhorn layers. AMC-Net~\cite{AMC-Net} modulates the weights of appearance and motion features and suppresses the redundant and misleading information by evaluating the importance of each modality. As each frame is separately processed in these two-stream methods, only an RGB image and an optical flow map are required to infer each frame. However, since they are highly dependent on the quality of optical flow maps (susceptible to low-quality optical flow maps), stable and reliable prediction cannot be achieved.

\vspace{1mm}
\noindent\textbf{Network learning strategy.} As the amount of training data for unsupervised VOS is not sufficient compared to that for other vision tasks, existing methods have adopted various network learning strategies for effective network training. AGS~\cite{AGS} exploits two image SOD datasets, DUT~\cite{DUT} and PASCAL-S~\cite{PASCAL-S}, both of which provide static gaze data and segmentation annotations for use as external data. COSNet and AGNN use pre-trained segmentation models that are trained on MSRA10K~\cite{MSRA10K} and DUT, and fine-tune the networks on the DAVIS 2016~\cite{DAVIS} training set. RTNet and FSNet pre-train the appearance stream on DUTS~\cite{DUTS} before the main training on videos. IMP~\cite{IMP} adopts a pre-trained semi-supervised VOS model trained on large datasets, such as COCO~\cite{COCO} and YouTube-VOS 2018~\cite{YTVOS}.

\section{Approach}
\subsection{Problem Formulation}
Unsupervised VOS aims to predict binary segmentation masks $O:=\{O^0, O^1, ..., O^{L-1}\}$ using input RGB images $I:=\{I^0, I^1, ..., I^{L-1}\}$, where L is the length of a given video sequence. To leverage motion as well as appearance information, optical flow maps $F:=\{F^0, F^1, ..., F^{L-1}\}$ are generated using a pre-trained optical flow model and are used as input after transforming 2-channel motion vectors to 3-channel RGB values. Following existing two-stream VOS methods, such as MATNet~\cite{MATNet}, FSNet~\cite{FSNet}, and AMC-Net~\cite{AMC-Net}, an input video is processed frame-by-frame in our method. When inferring $O^i$, only $I^i$ and $F^i$ are required.

\subsection{Motion-as-Option Network}
To reduce motion dependency of existing two-stream methods, which are susceptible to low-quality optical flow maps, we propose a novel motion-as-option network that flexibly leverages motion cues to reduce motion dependency. Figure~\ref{figure2} displays the architecture of our proposed motion-as-option network.

\vspace{1mm}
\noindent\textbf{Separate encoders.} Considering that appearance and motion cues have different advantages that can complement each other, existing two-stream methods employ strongly-connected encoders. After every encoding block, appearance and motion features exchange their information, imposing constraints on each other. As the appearance and motion encoders cannot be separated, they can be regarded as a single encoder with an RGB image and an optical flow map as its input. Therefore, existing two-stream VOS methods are very dependent on motion cues extracted from optical flow maps, which results in critical errors when low-quality optical flow maps are used as input.

Unlike the existing two-stream methods, we use two separate encoders to embed appearance features and motion features independently. Let us denote appearance features as $\{A_k\}_{k=1}^K$ and motion features as $\{M_k\}_{k=1}^K$, where $K$ is the number of blocks in the encoder and a higher $k$ value indicates higher-level features. The appearance features are extracted from RGB images, whereas the motion features are obtained from either RGB images or optical flow maps. After blending appearance features and motion features, fused features $\{X_k\}_{k=1}^K$ can be defined as
\begin{align}
&X_k = A_k + M_k~.
\label{eq1}
\end{align}
As appearance features and motion features are separately embedded and then fused, the separation of appearance and motion cues can be easily achieved. This makes our network robust against inaccurate motion cues, as the network becomes less dependent on the motion stream. Additionally, considering both RGB images and optical flow maps are used as input of the motion encoder during network training, network overfitting to explicit motion cues can be prevented. This also greatly increases usability and applicability of the motion-as-option network as it does not necessarily need optical flow maps during inference.

\vspace{1mm}
\noindent\textbf{Decoder.} The decoder refines the fused features and generates binary segmentation masks. To yield high-resolution masks, the fused features are gradually refined using decoding blocks that are designed similar to those of TBD~\cite{TBD}. Each decoding block comprises a convolutional layer that blends different features, a CBAM layer~\cite{CBAM} that reinforces both channel and spatial feature representations, and an upsampling layer that increases the spatial size of the features. Assuming $\Psi_k$ indicates the $k$-th decoding block, where a higher $k$ value indicates higher resolution, decoded features $\{D_k\}_{k=1}^K$ can be obtained as
\begin{align}
&D_k = 
\begin{cases}
\Psi_k(X_{K-k+1}) &k=1\\ \Psi_k(D_{k-1} \oplus X_{K-k+1}) &\text{otherwise}~,
\end{cases}
\end{align}
where $\oplus$ indicates channel concatenation. After $K$ decoding blocks, the final segmentation mask $O$ can be defined after value quantization using an argmax operation. The first and second channels denote the background and foreground segmentation maps, respectively.

\subsection{Collaborative Learning Strategy}
\label{cls}
The key property of the proposed motion-as-option network is that it optionally leverages optical flow maps. To train the network in accordance with our network design goals, training samples with and without optical flow maps are required during the network training. A simple approach for accomplishing this is intentionally and randomly discarding optical flow maps of VOS training samples. However, instead, we adopt SOD training samples to fully exploit the advantage of the proposed network that optical flow maps are not always required. When VOS samples are adopted, optical flow maps are used as input of the motion encoder, whereas when SOD samples are adopted, RGB images are used as input of the motion encoder.

Although providing training samples in a collaborative manner sounds reasonable as it maximizes the efficacy of the proposed network, it cannot be simply implemented on GPU devices due to the different formats of VOS and SOD data. To enable batch training to accelerate network training and ensure stability, we use a simple indexing trick. First, training data samples are generated from a dataset comprising VOS and SOD training samples. For each VOS training sample, an RGB image, an optical flow map, and a ground truth segmentation mask are loaded. In contrast, for each SOD training sample, an RGB image and a ground truth segmentation mask are loaded. Then, to ensure that the samples have the same format, void tensors are generated and regarded as optical flow maps for the SOD samples. To verify the validity of optical flow maps, a motion validity index is allocated for each training sample. For VOS and SOD samples, it is set to 1 and 0, respectively. Using the generated indices, feature fusion process is implemented as shown in Code Listing~\ref{listing1}.

\begin{python}[t!]
# A: appearance features
# M: motion features
# X: fused features
A, M, X = {}, {}, {}

# image:(Bx3xHxW), RGB images
# extract appearance features
a = image
for k in K:
    A[k] = app_block_k(a)
    a = A[k]

# i:(Bx1x1x1), motion validity indices
# flow:(Bx3xHxW), optical flow maps
# extract motion features
m = i * flow + (1 - i) * image
for k in K:
    M[k] = mo_block_k(m)
    m = M[k]

# generate fused features
for k in K:
    X[k] = A[k] + M[k]
\end{python}
\vspace{-2mm}
\begin{lstlisting}[frame=none, caption={Feature fusion process for batch training.}, label={listing1}]
\end{lstlisting}

\subsection{Implementation Details}
\noindent\textbf{Optical flow map.} To leverage motion information as well as appearance information, we extract semantic motion cues from optical flow maps. To generate the flow map at frame $i$, we consider frame $i$ as the starting frame and frame $i+1$ as the target frame. If $i$ is the last frame of a video, i.e., $L-1$, we regard $i-1$ as the target frame. As the optical flow estimation network, we use RAFT~\cite{RAFT} pre-trained on the Sintel~\cite{Sintel} dataset. Optical flow map generation is performed without changing the original resolution of the VOS data samples. To reduce redundant training and testing times, we generate and save the optical flow maps in advance, instead of calculating them immediately in the middle of inference.

\vspace{1mm}
\noindent\textbf{Encoder.} Following most existing approaches for unsupervised VOS, such as COSNet~\cite{COSNet}, AD-Net~\cite{AD-Net}, and MATNet~\cite{MATNet}, we adopt ResNet-101~\cite{resnet} as our backbone encoder. It comprises total four blocks, i.e., $K=4$. The features extracted from the $k$-th block have the scale of $1/2^{k+1}$ compared to the input resolution. To preserve rich feature representations learned from the large number of training samples, we initialize both appearance and motion encoders with ImageNet~\cite{imagenet} pre-trained version.

\subsection{Network Training}
\label{training}
\noindent\textbf{Data preparation.} To obtain training data diversity, we adopt both VOS and SOD datasets, as described in Section~\ref{cls}. As the VOS training dataset, DAVIS 2016~\cite{DAVIS} training set is employed. Note that FBMS~\cite{FBMS} also has a training set but is not adopted for network training following the common protocol in unsupervised VOS. As the SOD training dataset, DUTS~\cite{DUTS} is adopted. We use both DUTS training and test sets as our training dataset. The training data are randomly sampled from VOS and SOD samples with fixed probabilities of 25\% and 75\%, respectively.

\vspace{1mm}
\noindent\textbf{Training details.} During all the training stages, we resize the RGB images, optical flow maps, and segmentation masks to 384$\times$384 resolution. We use bicubic interpolation to resize RGB images and optical flow maps, while the nearest interpolation is used for segmentation masks to sustain the values to either 0 or 1. For network optimization, we use cross-entropy loss and the Adam optimizer~\cite{adam}. The learning rate is set to 1e-5 without learning the rate decay, and the batch size is set to 16. Following existing semi-supervised VOS algorithms, such as STM~\cite{STM}, KMN~\cite{KMN}, CFBI~\cite{CFBI}, and BMVOS~\cite{BMVOS}, we freeze all batch normalization layers~\cite{batchnorm} during training. Network training is implemented on two GeForce RTX 2080 Ti GPUs and takes less than 20 hours.

\begin{table*}
	\centering 
	\caption{Quantitative evaluation on the DAVIS 2016 validation set and FBMS test set. OF and PP indicate the use of optical flow estimation models and post-processing techniques, respectively.}
	\vspace{2mm}
	\small
	\begin{tabular}{p{2.3cm}P{2cm}P{1.5cm}P{0.5cm}P{0.5cm}P{1cm}P{1cm}P{1cm}P{1cm}P{1cm}}
		\toprule
		\multicolumn{6}{c}{} &\multicolumn{3}{c}{DAVIS 2016} &\multicolumn{1}{c}{FBMS}\\
		\cline{7-10}
		Method &Publication &Resolution &OF &PP &fps &$\mathcal{G}_\mathcal{M}$ &$\mathcal{J}_\mathcal{M}$ &$\mathcal{F}_\mathcal{M}$ &$\mathcal{J}_\mathcal{M}$\\
		\midrule
		PDB~\cite{PDB} &ECCV'18 &473$\times$473 & &\checkmark &20.0 &75.9 &77.2 &74.5 &74.0\\
		MOTAdapt~\cite{MOTAdapt} &ICRA'19 &- & &\checkmark &- &77.3 &77.2 &77.4 &-\\
		AGS~\cite{AGS} &CVPR'19 &473$\times$473 & &\checkmark &10.0 &78.6 &79.7 &77.4 &-\\
		COSNet~\cite{COSNet} &CVPR'19 &473$\times$473 & &\checkmark &- &80.0 &80.5 &79.4 &75.6\\
		AD-Net~\cite{AD-Net} &ICCV'19 &480$\times$854 & &\checkmark &4.00 &81.1 &81.7 &80.5 &-\\
		AGNN~\cite{AGNN} &ICCV'19 &473$\times$473 & &\checkmark &3.57 &79.9 &80.7 &79.1 &-\\
		MATNet~\cite{MATNet} &AAAI'20 &473$\times$473 &\checkmark &\checkmark &20.0 &81.6 &72.4 &80.7 &76.1\\
		WCS-Net~\cite{WCS-Net} &ECCV'20 &320$\times$320 & & &\underline{33.3} &81.5 &82.2 &80.7 &-\\
		DFNet~\cite{DFNet} &ECCV'20 &- & &\checkmark &3.57 &82.6 &83.4 &81.8 &-\\
		3DC-Seg~\cite{3DC-Seg} &BMVC'20 &480$\times$854 & &\checkmark &4.55 &84.5 &84.3 &84.7 &-\\
        F2Net~\cite{F2Net} &AAAI'21 &473$\times$473 & & &10.0 &83.7 &83.1 &84.4 &77.5\\
        RTNet~\cite{RTNet} &CVPR'21 &384$\times$672 &\checkmark &\checkmark &- &85.2 &\textbf{85.6} &84.7 &-\\
        FSNet~\cite{FSNet} &ICCV'21 &352$\times$352 &\checkmark &\checkmark &12.5 &83.3 &83.4 &83.1 &-\\
        TransportNet~\cite{TransportNet} &ICCV'21 &512$\times$512 &\checkmark & &12.5 &84.8 &84.5 &85.0 &\underline{78.7}\\
        AMC-Net~\cite{AMC-Net} &ICCV'21 &384$\times$384 &\checkmark &\checkmark &17.5 &84.6 &84.5 &84.6 &76.5\\
        D$^2$Conv3D~\cite{D^2Conv3D} &WACV'22 &480$\times$854 & & &- &\underline{86.0} &\underline{85.5} &86.5 &-\\
        IMP~\cite{IMP} &AAAI'22 &- & & &1.79 &85.6 &84.5 &\textbf{86.7} &77.5\\
		\midrule
		\textbf{TMO} & &384$\times$384 &\checkmark & &\textbf{43.2} &\textbf{86.1} &\textbf{85.6} &\underline{86.6} &\textbf{79.9}\\
		\bottomrule
	\end{tabular}
	\label{Table:DAVIS,FBMS}
\end{table*}

\begin{table*}
	\centering 
	\caption{Quantitative evaluation on the YouTube-Objects dataset. Performance is reported with the $\mathcal{J}$ mean.}
	\vspace{2mm}
	\small
	\begin{tabular}{p{1.9cm}P{1.2cm}P{0.9cm}P{0.9cm}P{0.9cm}P{0.9cm}P{0.9cm}P{0.9cm}P{0.9cm}P{1.2cm}P{0.9cm}|P{0.9cm}}
		\toprule
		Method &Aeroplane &Bird &Boat &Car &Cat &Cow &Dog &Horse &Motorbike &Train &Mean\\
		\midrule
		PDB~\cite{PDB} &78.0 &80.0 &58.9 &76.5 &63.0 &64.1 &70.1 &\underline{67.6} &58.4 &35.3 &65.5\\
		AGS~\cite{AGS} &\textbf{87.7} &76.7 &\textbf{72.2} &78.6 &69.2 &64.6 &73.3 &64.4 &62.1 &48.2 &69.7\\
		COSNet~\cite{COSNet} &81.1 &75.7 &\underline{71.3} &77.6 &66.5 &69.8 &\underline{76.8} &67.4 &67.7 &46.8 &70.5\\
		AGNN~\cite{AGNN} &71.1 &75.9 &70.7 &78.1 &67.9 &69.7 &\textbf{77.4} &67.3 &\underline{68.3} &47.8 &70.8\\
		MATNet~\cite{MATNet} &72.9 &77.5 &66.9 &79.0 &\textbf{73.7} &67.4 &75.9 &63.2 &62.6 &51.0 &69.0\\
		WCS-Net~\cite{WCS-Net} &81.8 &\textbf{81.1} &67.7 &79.2 &64.7 &65.8 &73.4 &\textbf{68.6} &\textbf{69.7} &49.2 &70.5\\
		RTNet~\cite{RTNet} &84.1 &80.2 &70.1 &\underline{79.5} &71.8 &\underline{70.1} &71.3 &65.1 &64.6 &\underline{53.3} &71.0\\
		AMC-Net~\cite{AMC-Net} &78.9 &\underline{80.9} &67.4 &\textbf{82.0} &69.0 &69.6 &75.8 &63.0 &63.4 &\textbf{57.8} &\underline{71.1}\\
		\midrule
		\textbf{TMO} &\underline{85.7} &80.0 &70.1 &78.0 &\underline{73.6} &\textbf{70.3} &\underline{76.8} &66.2 &58.6 &47.0 &\textbf{71.5}\\
		\bottomrule
	\end{tabular}
	\label{Table:YTOBJ}
\end{table*}

\section{Experiments}
In this section, we first describe the datasets used in this study in Section~\ref{dataset}. Evaluation metrics for quantitatively evaluating the method performance are described in Section~\ref{metric}. The quantitative and qualitative comparison of our approach to other state-of-the-art methods is presented in Section~\ref{quantitative} and Section~\ref{qualitative}, respectively. Finally, we thoroughly validate the effectiveness of our proposed approach by conducting an extensive analysis in Section~\ref{analysis}. Our method is abbreviated as TMO.

\subsection{Datasets}
\label{dataset}
To validate the effectiveness of our proposed approach, we use four datasets that are widely adopted for unsupervised VOS. For network training, DUTS~\cite{DUTS} and DAVIS 2016~\cite{DAVIS} are used. For network testing, DAVIS 2016, FBMS~\cite{FBMS}, and YouTube-Objects~\cite{YTOBJ} are used.

\vspace{1mm}
\noindent\textbf{DUTS.} DUTS is the largest SOD dataset, comprising 10,553 training images and 5,019 test images, each with densely-annotated ground truth segmentation masks.

\vspace{1mm}
\noindent\textbf{DAVIS 2016.} DAVIS 2016 is one of the most popular datasets for VOS tasks. It contains 30 training video sequences and 20 validation video sequences that only contain single-object scenarios. 

\vspace{1mm}
\noindent\textbf{FBMS.} FBMS comprises 59 video sequences, and a total of 720 frames are annotated. In some sequences, multiple objects are annotated as salient objects.

\vspace{1mm}
\noindent\textbf{YouTube-Objects.} YouTube-Objects dataset constitutes videos collected from YouTube and contains 10 object classes. Each class contains between 9 and 24 video sequences. As there is no training set / test set separation, it is only used for network validation.

\subsection{Evaluation Metrics}
\label{metric}
The performance of the unsupervised VOS methods can be quantitatively evaluated using a protocol similar to that for the general image segmentation tasks. Usually, two kinds of measurements are adopted: $\mathcal{J}$ and $\mathcal{F}$. $\mathcal{J}$ is a metric that measures the region accuracy. It is equal to a normal intersection-over-union (IoU) metric, which can be represented as
\begin{align}
&\mathcal{J} = \left| \frac{M_{gt} \cap M_{pred}}{M_{gt} \cup M_{pred}} \right|~,
\label{eq3}
\end{align}
where $M_{gt}$ and $M_{pred}$ denote the ground truth and predicted binary segmentation masks, respectively. Contour accuracy $\mathcal{F}$ is a metric that is also based on IoU, but it is calculated only for the object boundaries as
\begin{align}
&\text{Precision} = \left| \frac{M_{gt} \cap M_{pred}}{M_{pred}} \right|~,
\end{align}
\begin{align}
&\text{Recall} = \left| \frac{M_{gt} \cap M_{pred}}{M_{gt}} \right|~,
\end{align}
\begin{align}
&\mathcal{F} = \frac{2 \times \text{Precision} \times \text{Recall}}{\text{Precision} + \text{Recall}}~.
\end{align}
$\mathcal{G}$ metric, which is the average of $\mathcal{J}$ and $\mathcal{F}$, is also a widely-used metric for evaluating VOS performance.

\begin{figure*}[t]
	\centering
	\includegraphics[width=1\linewidth]{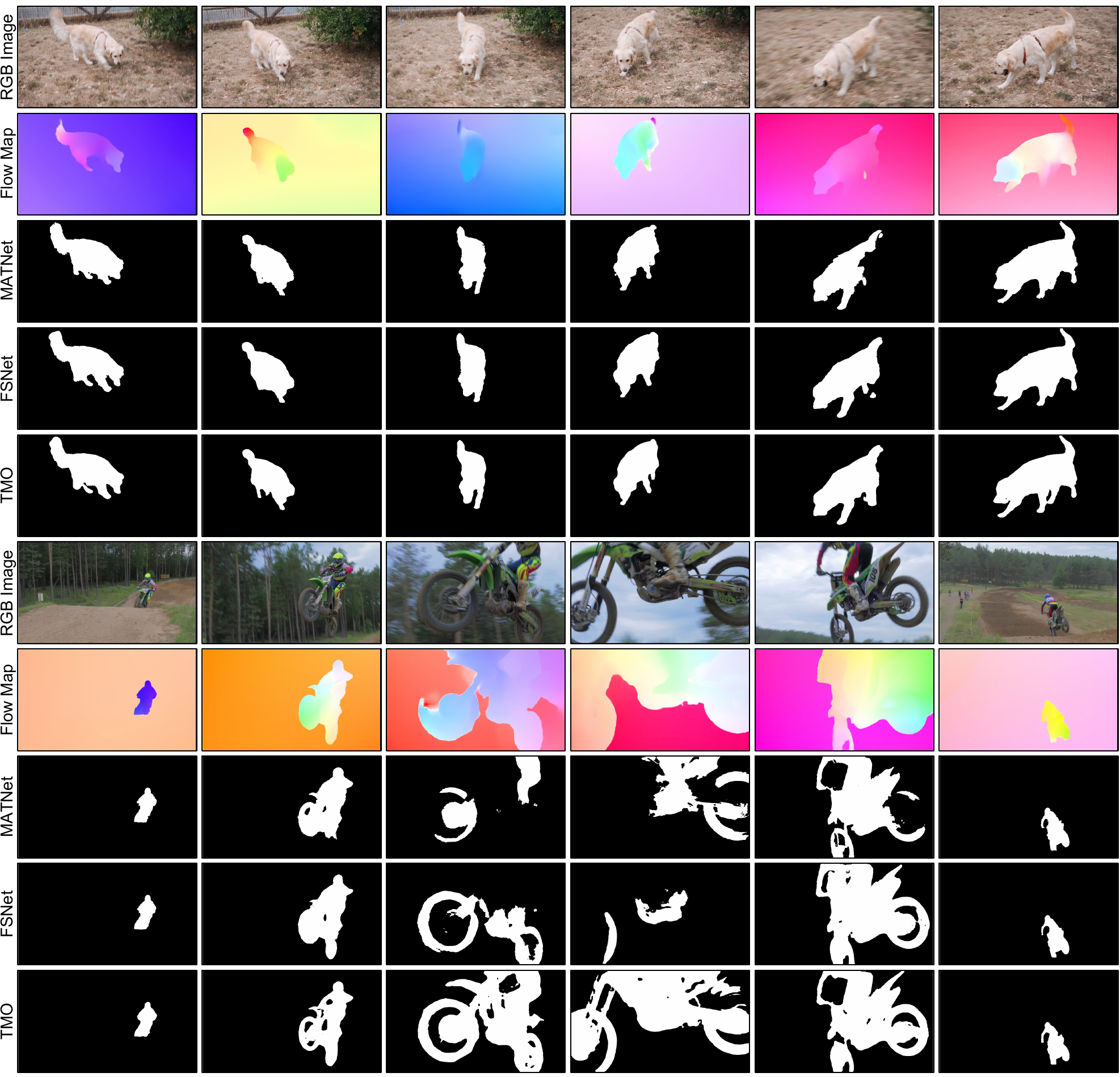}
	\caption{Qualitative comparison of different methods on the DAVIS 2016 validation set.}
	\label{figure3}
\end{figure*}

\subsection{Quantitative Results}
\label{quantitative}
We quantitatively evaluate the performance of our method and compare it with that of other state-of-the-art methods on three popular benchmark datasets for unsupervised VOS: DAVIS 2016~\cite{DAVIS} validation set, FBMS~\cite{FBMS} test set, and YouTube-Objects~\cite{YTOBJ} dataset.

\vspace{1mm}
\noindent\textbf{DAVIS 2016.} In Table~\ref{Table:DAVIS,FBMS}, we evaluate the performance of our method on the DAVIS 2016 validation set. For a fair comparison, we also report the use of post-processing techniques, such as fully connected CRF~\cite{densecrf} and instance pruning proposed in AD-Net~\cite{AD-Net}, which have heavy computational burden. In addition, to consider the efficiency of each method, the inference time is reported for the methods whose fps is publicly available. Note that both the post-processing time and optical flow map generation time are not considered. The table shows that D$^2$Conv3D~\cite{D^2Conv3D} achieves a notable performance with a $\mathcal{G}$ score of 86.0\%, but as it is based on 3D convolutional layers, it requires considerable network pre-training data, such as Kinetics400~\cite{kinetics} and Sports-1M~\cite{sports}. IMP~\cite{IMP} also exhibits remarkable performance with a $\mathcal{G}$ score of 85.6\%, but as it requires global observation on all the frames of a video to select the optimal starting frame, it cannot run in an online manner. Unlike the methods that sacrifice applicability to obtain higher accuracy, WCS-Net~\cite{WCS-Net} is quite practical with a relatively-fast inference speed of 33.3 fps. However, its performance is not comparable to those of other state-of-the-art methods. Even without post-processing techniques and maintaining online availability, our method outperforms all other methods with a $\mathcal{G}$ score of 86.1\% and an exceptionally-fast inference speed of 43.2 fps.

\vspace{1mm}
\noindent\textbf{FBMS.} Quantitative evaluation on the FBMS test set is also presented in Table~\ref{Table:DAVIS,FBMS}. Among the existing methods, TransportNet~\cite{TransportNet} achieves the best performance with a $\mathcal{J}$ score of 78.7\%. Performances of F2Net~\cite{F2Net} and IMP are also impressive, showing a $\mathcal{J}$ score of 77.5\%. Our method surpasses all other methods with a $\mathcal{J}$ score of 79.9\%.

\vspace{1mm}
\noindent\textbf{YouTube-Objects.} In Table~\ref{Table:YTOBJ}, performance of the state-of-the-art methods on the YouTube-Objects dataset is quantitatively compared. For the YouTube-Objects dataset, the model performance is evaluated using class accuracy, which denotes the mean score of the sequences in each class, and overall accuracy, which denotes the mean score of all sequences in the entire dataset. TMO achieves the highest overall accuracy of 71.5\% $\mathcal{J}$ score. The performance on the YouTube-Objects dataset demonstrates the effectiveness of our approach for challenging scenarios.

\subsection{Qualitative Results}
\label{qualitative}
We qualitatively compare our method to state-of-the-art MATNet~\cite{MATNet} and FSNet~\cite{FSNet} on the DAVIS 2016~\cite{DAVIS} validation set in Figure~\ref{figure3}. For a fair comparison, we select the methods that leverage motion cues obtained from pre-trained optical flow estimation models as our proposed method. As shown in the figure, MATNet and FSNet generate noisy results, especially when optical flow maps are confusing or unclear, e.g., the third and fourth frames in the lower sequence. In such cases, they cannot appropriately deal with the error of the flow maps, as the network is highly dependent on motion cues. Unlike those methods, TMO can consistently generate fine and accurate segmentation masks even when the quality of the flow maps is low, as the entire pipeline is learned by treating motion as option. The qualitative results also demonstrate the superiority of our approach for reliable and stable VOS compared to other state-of-the-art VOS approaches.

\subsection{Analysis}
\label{analysis}
We conduct an ablation study on the network training and testing protocols of our method in Table~\ref{Table:ablation}. Networks trained with VOS samples, sequentially trained with SOD and VOS samples, and collaboratively trained with VOS and SOD samples are compared with and without the use of the motion stream. As described in Section~\ref{training}, we use DAVIS 2016~\cite{DAVIS} training set as VOS samples and DUTS~\cite{DUTS} training set and test set as SOD samples. For network testing, $\mathcal{G}$ score is adopted for the DAVIS 2016 validation set, whereas for the FBMS~\cite{FBMS} test set and YouTube-Objects~\cite{YTOBJ} dataset, $\mathcal{J}$ score is adopted.

\vspace{1mm}
\noindent\textbf{Collaborative learning strategy.} We validate the effectiveness of our proposed collaborative network learning strategy by comparing model~\Romannum{1}, model~\Romannum{2}, and model~\Romannum{3}. If only the VOS sampels are adopted for network training, as in model~\Romannum{1}, the performance is unsatisfactory due to the small number of training samples in the DAVIS 2016 training set. The low performance on all datasets shows that network overfitting occurs. To prevent the overfitting problem, existing two-stream methods, including COSNet~\cite{COSNet} and FSNet~\cite{FSNet}, pre-train the network on SOD datasets, such as MSRA10K~\cite{MSRA10K} and DUTS datasets, and then fine-tune it on the DAVIS 2016 training set. To simulate this protocol for a fair comparison, we first pre-train our network on the SOD samples and then fine-tune it on the VOS samples, as in model~\Romannum{2}. Although this method can relieve the overfitting problem to some extent, the network is still overly fitted to the training set. In contrast, if our proposed collaborative network learning strategy is adopted, as in model~\Romannum{3}, the model does not suffer from the overfitting problem as both VOS and SOD samples are jointly used for network training. On all testing datasets, our approach outperforms other training protocols, demonstrating the effectiveness of a collaborative network learning strategy.

\vspace{1mm}
\noindent\textbf{Motion dependency.} To determine to what extent each training strategy makes a model dependent on motion cues, we also report the performance when RGB images are fed into the motion encoder. For model versions trained with the VOS samples and sequentially trained with the SOD and VOS samples, performance degradation when optical flow maps are not available is substantial for the DAVIS 2016 validation set. However if the collaborative learning strategy is adopted, the performance degradation is only 6.1\%, verifying that the collaborative learning strategy greatly reduces motion dependency of the network. Notably, compared to the DAVIS 2016 validation set, the FBMS test set and YouTube-Objects dataset seem to not be considerably dependent on motion cues. Specifically, model~\Romannum{6} yields a higher score than model~\Romannum{3} on the FBMS test set and YouTube-Objects dataset. This also supports the need to treat motion as option for reducing motion dependency since optical flow maps are not always reliable.

\begin{table}[t!]
	\centering 
	\caption{Ablation study on network training and testing protocols. Training indicates which training protocols are adopted for network training. Testing indicates input of the motion encoder during inference. D, F, and Y denote DAVIS 2016 validation set, FBMS test set, and YouTube-Objects dataset, respectively.}
	\vspace{2mm}
	\small
	\begin{tabular}{c|c|c|P{0.7cm}P{0.7cm}P{0.7cm}}
		\toprule
		Version &Training &Testing &D &F &Y\\
		\midrule
		\Romannum{1} &VOS &Flow &76.5 &59.2 &57.2\\
		\Romannum{2} &SOD $\rightarrow$ VOS &Flow &82.1 &74.7 &63.0\\
        \Romannum{3} &VOS~\&~SOD &Flow &86.1 &79.9 &71.5\\
        \midrule
		\Romannum{4} &VOS &Image &63.7 &52.8 &52.6\\
		\Romannum{5} &SOD $\rightarrow$ VOS &Image &73.0 &76.8 &69.3\\
        \Romannum{6} &VOS~\&~SOD &Image &80.0 &80.0 &73.1\\
		\bottomrule
	\end{tabular}
	\label{Table:ablation}
\end{table}

\section{Conclusion}
In unsupervised VOS, the combined use of appearance and motion streams has been an effective and powerful tool. However, as existing two-stream methods fuse appearance and motion cues in early stage, they are very dependent on motion cues. This makes them susceptible to low-quality optical flow maps, degrading their usability and reliability. To relieve this concern, we propose a motion-as-option network that is not highly dependent on motion cues and can also operate only with appearance cues. Additionally, to fully leverage this property, a collaborative network learning strategy is proposed. On all the public benchmark datasets, our approach affords a new state-of-the-art performance with real-time inference speed. We believe that our simple, fast, and strong approach can serve as a solid baseline for future VOS research.

\vspace*{\fill}
\noindent\textbf{Acknowledgement.} This work was supported by the Institute of Information \& communications Technology Planning \& Evaluation (IITP) grant funded by the Korea government (MSIT) (No. 2021-0-00172, The development~of human Re-identification and masked face recognition based on CCTV camera) and the KIST Institutional Program (Project No.2E31051-21-203).

{\small
\bibliographystyle{ieee_fullname}
\bibliography{egbib}

\begin{thebibliography}{10}\itemsep=-1pt

\bibitem{Sintel}
D.~J. Butler, J. Wulff, G.~B. Stanley, and M.~J. Black.
\newblock A naturalistic open source movie for optical flow evaluation.
\newblock In {A. Fitzgibbon et al. (Eds.)}, editor, {\em European Conf. on
  Computer Vision (ECCV)}, Part IV, LNCS 7577, pages 611--625. Springer-Verlag,
  Oct. 2012.

\bibitem{MSRA10K}
Ming-Ming Cheng, Niloy~J. Mitra, Xiaolei Huang, Philip H.~S. Torr, and Shi-Min
  Hu.
\newblock Global contrast based salient region detection.
\newblock {\em IEEE TPAMI}, 37(3):569--582, 2015.

\bibitem{BMVOS}
Suhwan Cho, Heansung Lee, Minjung Kim, Sungjun Jang, and Sangyoun Lee.
\newblock Pixel-level bijective matching for video object segmentation.
\newblock In {\em Proceedings of the IEEE/CVF Winter Conference on Applications
  of Computer Vision}, pages 129--138, 2022.

\bibitem{TBD}
Suhwan Cho, Heansung Lee, Minhyeok Lee, Chaewon Park, Sungjun Jang, Minjung
  Kim, and Sangyoun Lee.
\newblock Tackling background distraction in video object segmentation.
\newblock {\em arXiv preprint arXiv:2207.06953}, 2022.

\bibitem{resnet}
Kaiming He, Xiangyu Zhang, Shaoqing Ren, and Jian Sun.
\newblock Deep residual learning for image recognition.
\newblock In {\em Proceedings of the IEEE conference on computer vision and
  pattern recognition}, pages 770--778, 2016.

\bibitem{batchnorm}
Sergey Ioffe and Christian Szegedy.
\newblock Batch normalization: Accelerating deep network training by reducing
  internal covariate shift.
\newblock {\em arXiv preprint arXiv:1502.03167}, 2015.

\bibitem{FSNet}
Ge-Peng Ji, Keren Fu, Zhe Wu, Deng-Ping Fan, Jianbing Shen, and Ling Shao.
\newblock Full-duplex strategy for video object segmentation.
\newblock In {\em Proceedings of the IEEE/CVF international conference on
  computer vision}, pages 4922--4933, 2021.

\bibitem{sports}
Andrej Karpathy, George Toderici, Sanketh Shetty, Thomas Leung, Rahul
  Sukthankar, and Li Fei-Fei.
\newblock Large-scale video classification with convolutional neural networks.
\newblock In {\em Proceedings of the IEEE conference on Computer Vision and
  Pattern Recognition}, pages 1725--1732, 2014.

\bibitem{kinetics}
Will Kay, Joao Carreira, Karen Simonyan, Brian Zhang, Chloe Hillier, Sudheendra
  Vijayanarasimhan, Fabio Viola, Tim Green, Trevor Back, Paul Natsev, et~al.
\newblock The kinetics human action video dataset.
\newblock {\em arXiv preprint arXiv:1705.06950}, 2017.

\bibitem{adam}
Diederik~P Kingma and Jimmy Ba.
\newblock Adam: A method for stochastic optimization.
\newblock {\em arXiv preprint arXiv:1412.6980}, 2014.

\bibitem{densecrf}
Philipp Kr{\"a}henb{\"u}hl and Vladlen Koltun.
\newblock Efficient inference in fully connected crfs with gaussian edge
  potentials.
\newblock {\em Advances in neural information processing systems}, 24, 2011.

\bibitem{imagenet}
Alex Krizhevsky, Ilya Sutskever, and Geoffrey~E Hinton.
\newblock Imagenet classification with deep convolutional neural networks.
\newblock {\em Communications of the ACM}, 60(6):84--90, 2017.

\bibitem{IMP}
Youngjo Lee, Hongje Seong, and Euntai Kim.
\newblock Iteratively selecting an easy reference frame makes unsupervised
  video object segmentation easier.
\newblock In {\em Proceedings of the AAAI Conference on Artificial
  Intelligence}, volume~36, pages 1245--1253, 2022.

\bibitem{PASCAL-S}
Yin Li, Xiaodi Hou, Christof Koch, James~M Rehg, and Alan~L Yuille.
\newblock The secrets of salient object segmentation.
\newblock In {\em Proceedings of the IEEE conference on computer vision and
  pattern recognition}, pages 280--287, 2014.

\bibitem{COCO}
Tsung-Yi Lin, Michael Maire, Serge Belongie, James Hays, Pietro Perona, Deva
  Ramanan, Piotr Doll{\'a}r, and C~Lawrence Zitnick.
\newblock Microsoft coco: Common objects in context.
\newblock In {\em European conference on computer vision}, pages 740--755.
  Springer, 2014.

\bibitem{F2Net}
Daizong Liu, Dongdong Yu, Changhu Wang, and Pan Zhou.
\newblock F2net: Learning to focus on the foreground for unsupervised video
  object segmentation.
\newblock In {\em Proceedings of the AAAI Conference on Artificial
  Intelligence}, volume~35, pages 2109--2117, 2021.

\bibitem{COSNet}
Xiankai Lu, Wenguan Wang, Chao Ma, Jianbing Shen, Ling Shao, and Fatih Porikli.
\newblock See more, know more: Unsupervised video object segmentation with
  co-attention siamese networks.
\newblock In {\em Proceedings of the IEEE/CVF conference on computer vision and
  pattern recognition}, pages 3623--3632, 2019.

\bibitem{3DC-Seg}
Sabarinath Mahadevan, Ali Athar, Aljo{\v{s}}a O{\v{s}}ep, Sebastian Hennen,
  Laura Leal-Taix{\'e}, and Bastian Leibe.
\newblock Making a case for 3d convolutions for object segmentation in videos.
\newblock {\em arXiv preprint arXiv:2008.11516}, 2020.

\bibitem{FBMS}
Peter Ochs, Jitendra Malik, and Thomas Brox.
\newblock Segmentation of moving objects by long term video analysis.
\newblock {\em IEEE transactions on pattern analysis and machine intelligence},
  36(6):1187--1200, 2013.

\bibitem{STM}
Seoung~Wug Oh, Joon-Young Lee, Ning Xu, and Seon~Joo Kim.
\newblock Video object segmentation using space-time memory networks.
\newblock In {\em Proceedings of the IEEE/CVF International Conference on
  Computer Vision}, pages 9226--9235, 2019.

\bibitem{DAVIS}
Federico Perazzi, Jordi Pont-Tuset, Brian McWilliams, Luc Van~Gool, Markus
  Gross, and Alexander Sorkine-Hornung.
\newblock A benchmark dataset and evaluation methodology for video object
  segmentation.
\newblock In {\em Proceedings of the IEEE Conference on Computer Vision and
  Pattern Recognition}, pages 724--732, 2016.

\bibitem{YTOBJ}
Alessandro Prest, Christian Leistner, Javier Civera, Cordelia Schmid, and
  Vittorio Ferrari.
\newblock Learning object class detectors from weakly annotated video.
\newblock In {\em 2012 IEEE Conference on computer vision and pattern
  recognition}, pages 3282--3289. IEEE, 2012.

\bibitem{RTNet}
Sucheng Ren, Wenxi Liu, Yongtuo Liu, Haoxin Chen, Guoqiang Han, and Shengfeng
  He.
\newblock Reciprocal transformations for unsupervised video object
  segmentation.
\newblock In {\em Proceedings of the IEEE/CVF conference on computer vision and
  pattern recognition}, pages 15455--15464, 2021.

\bibitem{D^2Conv3D}
Christian Schmidt, Ali Athar, Sabarinath Mahadevan, and Bastian Leibe.
\newblock D2conv3d: Dynamic dilated convolutions for object segmentation in
  videos.
\newblock In {\em Proceedings of the IEEE/CVF Winter Conference on Applications
  of Computer Vision}, pages 1200--1209, 2022.

\bibitem{KMN}
Hongje Seong, Junhyuk Hyun, and Euntai Kim.
\newblock Kernelized memory network for video object segmentation.
\newblock In {\em European Conference on Computer Vision}, pages 629--645.
  Springer, 2020.

\bibitem{MOTAdapt}
Mennatullah Siam, Chen Jiang, Steven Lu, Laura Petrich, Mahmoud Gamal, Mohamed
  Elhoseiny, and Martin Jagersand.
\newblock Video object segmentation using teacher-student adaptation in a human
  robot interaction (hri) setting.
\newblock In {\em 2019 International Conference on Robotics and Automation
  (ICRA)}, pages 50--56. IEEE, 2019.

\bibitem{PDB}
Hongmei Song, Wenguan Wang, Sanyuan Zhao, Jianbing Shen, and Kin-Man Lam.
\newblock Pyramid dilated deeper convlstm for video salient object detection.
\newblock In {\em Proceedings of the European conference on computer vision
  (ECCV)}, pages 715--731, 2018.

\bibitem{RAFT}
Zachary Teed and Jia Deng.
\newblock Raft: Recurrent all-pairs field transforms for optical flow.
\newblock In {\em European conference on computer vision}, pages 402--419.
  Springer, 2020.

\bibitem{DUTS}
Lijun Wang, Huchuan Lu, Yifan Wang, Mengyang Feng, Dong Wang, Baocai Yin, and
  Xiang Ruan.
\newblock Learning to detect salient objects with image-level supervision.
\newblock In {\em CVPR}, 2017.

\bibitem{AGNN}
Wenguan Wang, Xiankai Lu, Jianbing Shen, David~J Crandall, and Ling Shao.
\newblock Zero-shot video object segmentation via attentive graph neural
  networks.
\newblock In {\em Proceedings of the IEEE/CVF international conference on
  computer vision}, pages 9236--9245, 2019.

\bibitem{AGS}
Wenguan Wang, Hongmei Song, Shuyang Zhao, Jianbing Shen, Sanyuan Zhao,
  Steven~CH Hoi, and Haibin Ling.
\newblock Learning unsupervised video object segmentation through visual
  attention.
\newblock In {\em Proceedings of the IEEE/CVF Conference on Computer Vision and
  Pattern Recognition}, pages 3064--3074, 2019.

\bibitem{CBAM}
Sanghyun Woo, Jongchan Park, Joon-Young Lee, and In~So Kweon.
\newblock Cbam: Convolutional block attention module.
\newblock In {\em Proceedings of the European conference on computer vision
  (ECCV)}, pages 3--19, 2018.

\bibitem{YTVOS}
Ning Xu, Linjie Yang, Yuchen Fan, Dingcheng Yue, Yuchen Liang, Jianchao Yang,
  and Thomas Huang.
\newblock Youtube-vos: A large-scale video object segmentation benchmark.
\newblock {\em arXiv preprint arXiv:1809.03327}, 2018.

\bibitem{DUT}
Chuan Yang, Lihe Zhang, Huchuan Lu, Xiang Ruan, and Ming-Hsuan Yang.
\newblock Saliency detection via graph-based manifold ranking.
\newblock In {\em Proceedings of the IEEE conference on computer vision and
  pattern recognition}, pages 3166--3173, 2013.

\bibitem{AMC-Net}
Shu Yang, Lu Zhang, Jinqing Qi, Huchuan Lu, Shuo Wang, and Xiaoxing Zhang.
\newblock Learning motion-appearance co-attention for zero-shot video object
  segmentation.
\newblock In {\em Proceedings of the IEEE/CVF International Conference on
  Computer Vision}, pages 1564--1573, 2021.

\bibitem{AD-Net}
Zhao Yang, Qiang Wang, Luca Bertinetto, Weiming Hu, Song Bai, and Philip~HS
  Torr.
\newblock Anchor diffusion for unsupervised video object segmentation.
\newblock In {\em Proceedings of the IEEE/CVF International Conference on
  Computer Vision}, pages 931--940, 2019.

\bibitem{CFBI}
Zongxin Yang, Yunchao Wei, and Yi Yang.
\newblock Collaborative video object segmentation by foreground-background
  integration.
\newblock In {\em European Conference on Computer Vision}, pages 332--348.
  Springer, 2020.

\bibitem{TransportNet}
Kaihua Zhang, Zicheng Zhao, Dong Liu, Qingshan Liu, and Bo Liu.
\newblock Deep transport network for unsupervised video object segmentation.
\newblock In {\em Proceedings of the IEEE/CVF International Conference on
  Computer Vision}, pages 8781--8790, 2021.

\bibitem{WCS-Net}
Lu Zhang, Jianming Zhang, Zhe Lin, Radom{\'\i}r M{\v{e}}ch, Huchuan Lu, and You
  He.
\newblock Unsupervised video object segmentation with joint hotspot tracking.
\newblock In {\em European Conference on Computer Vision}, pages 490--506.
  Springer, 2020.

\bibitem{DFNet}
Mingmin Zhen, Shiwei Li, Lei Zhou, Jiaxiang Shang, Haoan Feng, Tian Fang, and
  Long Quan.
\newblock Learning discriminative feature with crf for unsupervised video
  object segmentation.
\newblock In {\em European Conference on Computer Vision}, pages 445--462.
  Springer, 2020.

\bibitem{MATNet}
Tianfei Zhou, Shunzhou Wang, Yi Zhou, Yazhou Yao, Jianwu Li, and Ling Shao.
\newblock Motion-attentive transition for zero-shot video object segmentation.
\newblock In {\em Proceedings of the AAAI Conference on Artificial
  Intelligence}, volume~34, pages 13066--13073, 2020.

\end{thebibliography}
}

\end{document}